\documentclass[10pt,twocolumn,letterpaper]{article}

\usepackage{iccv}
\usepackage{times}
\usepackage{epsfig}
\usepackage{graphicx,subfigure}
\usepackage{amsmath}
\usepackage{amssymb}
\usepackage{enumitem}
\usepackage{array}
\usepackage{longtable}
\usepackage{booktabs}
\usepackage{fancyhdr}
\usepackage{bbm}
\usepackage{multirow}
\usepackage{color}

\usepackage[breaklinks=true,bookmarks=false]{hyperref}

\iccvfinalcopy % *** Uncomment this line for the final submission

\def\iccvPaperID{1452} % *** Enter the ICCV Paper ID here

\ificcvfinal\fi

\begin{document}

%%%%%%%%% TITLE
\title{Attentional Feature-Pair Relation Networks for Accurate Face Recognition}

\author{Bong-Nam Kang${}^{1,3}$, Yonghyun Kim${}^{2,3}$, Bongjin Jun${}^{1}$, Daijin Kim${}^{3}$\\
	${}^{1}$StradVision, Inc. ~~
	${}^{2}$Kakao Corp. ~~
	%${}^{3}$Department of Computer Science and Engineering, POSTECH, Korea\\
	${}^{3}$POSTECH\\
	%{\tt\small \{bnkang, gkyh0805, dkim\}@postech.ac.kr}, ~~{\tt\small bongjin.jun@stradvision.com}
	{\tt\small \{bongnam.kang, bongjin.jun\}@stradvision.com}, ~~{\tt\small aiden.kyh@kakaocorp.com}, ~~{\tt\small dkim@postech.ac.kr}
}

\maketitle
\ificcvfinal\thispagestyle{empty}\fi

%%%%%%%%% ABSTRACT
\begin{abstract}
   Human face recognition is one of the most important research areas in biometrics. However, the robust face recognition under a drastic change of the facial pose, expression, and illumination is a big challenging problem for its practical application. Such variations make face recognition more difficult.
   In this paper, we propose a novel face recognition method, called Attentional Feature-pair Relation Network (AFRN), which represents the face by the relevant pairs of local appearance block features with their attention scores.
   The AFRN represents the face by all possible pairs of the 9$\times$9 local appearance block features, the importance of each pair is considered by the attention map that is obtained from the low-rank bilinear pooling, and each pair is weighted by its corresponding attention score. %feature-pair bilinear attention or network
   To increase the accuracy, we select top-$K$ pairs of local appearance block features as relevant facial information and drop the remaining irrelevant. The weighted top-$K$ pairs are propagated to extract the joint feature-pair relation by using bilinear attention network.     
   %   In experiments, we show the effectiveness of the proposed AFRN and achieve the outstanding performance in the 1:1 face verification and 1:N face identification tasks compared to existing state-of-the-art methods on the challenging Labeled Faces in the Wild (LFW), YouTube Faces (YTF), IARPA Janus Benchmark A (IJB-A), and IARPA Janus Benchmark B (IJB-B) datasets.
   In experiments, we show the effectiveness of the proposed AFRN and achieve the outstanding performance in the 1:1 face verification and 1:N face identification tasks compared to existing state-of-the-art methods on the challenging LFW, YTF, CALFW, CPLFW, CFP, AgeDB, IJB-A, IJB-B, and IJB-C datasets.
\end{abstract}

%%%%%%%%% BODY TEXT
\section{Introduction}\label{sec:introduction}
Face recognition is one of the most important and interesting research areas in biometrics. However, the human appearances would be drastically changed under the unconstrained environment and the intra-person variations could overwhelm the inter-person variations, which make the face recognition difficult. Therefore, better face recognition requires for reducing the intra-person variations while enlarging the inter-person differences under the unconstrained environment.

\begin{figure*}[t]
	\centering
	\includegraphics[width=0.9\textwidth]{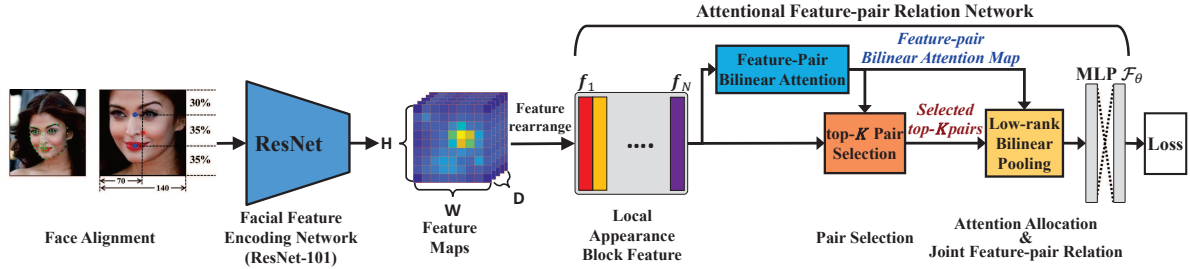}
	\caption{Working principle of the proposed Attentional Feature-pair Relation Network.}
	\label{fig:fig1}
\end{figure*}

Recent studies have targeted the same goal that minimizes the inter-person variations and maximizes the intra-person variations, either explicitly or implicitly. 
In deep learning-based face recognition methods, the deeply learned and embedded features are required to be not only separable but also discriminative to classify face images among different identities. This implies that the representation of a certain person \textit{A} stays unchanged regardless of who he/she is compared with, and it has to be discriminative enough to distinguish \textit{A} from all other persons. 
Chen \textit{et al.} achieved good recognition performance \cite{FaceVerification_WACV2016} by extracting feature representations via the CNN. And then, those features are applied to learn metric matrix to project the feature vector into a low-dimensional space in order to maximize the between-class variation and minimize within-class variation via the joint Bayesian metric learning.
Chowdhury \textit{et al.} applied the bilinear CNN architecture \cite{B-CNN_WACV2016} to the face identification task. 
Hassner \textit{et al.} proposed the pooling faces \cite{PoolingFace_CVPRW2016} that aligned faces in the 3D and binned them according to head pose and image quality. 
Masi \textit{et al.} proposed the pose-aware models (PAMs) \cite{POSE_AWARE_CVPR2016} that handled pose variability by learning pose-aware models for frontal, half-profile, and full-profile poses to improve face recognition performance in an unconstrained environment.
Sankaranarayanan \textit{et al.} \cite{Triplet_Prabablistic_Embedding_FR} proposed the triplet probabilistic embedding (TPE) that coupled a CNN-based approach with a low-dimensional discriminative embedding learned using triplet probability constraints. 
Crosswhite \textit{et al.} proposed the template adaptation (TA) \cite{Template_ADAPT_FGR2017} that was a form of transfer learning to the set of media in a template, which obtained better performance than the TPE on the IJB-A dataset by combining the CNN features with template adaptation. 
Yang \textit{et al.} proposed the neural aggregation network (NAN) \cite{NAN_CVPR2017} that produced a compact and fixed dimension feature representation. It adaptively aggregated the features to form a single feature inside the convex hull spanned by them and learned to advocate high-quality face images while repelling low-quality face images such as blurred, occluded and improperly exposed faces.
Ranjan \textit{et al.} \cite{L2_Softmax_FR_2017} added an L2-constraint to the feature descriptors which restricted them to lie on a hypersphere of a fixed radius, where minimizing the softmax loss is equivalent to maximizing the cosine similarity for the positive pairs and minimizing it for the negative pairs.
However, the above mentioned methods extracted the holistic features and did not designate what parts of the feature are meaningful and what parts of the features are separable and discriminative. Therefore, it is difficult to know what kind of features are used to discriminate the identities of face images clearly.

To overcome this disadvantage, some research efforts have been made regarding to the facial part-based representations for face recognition.
In DeepID \cite{DeepID} and DeepID2 \cite{DeepID2}, a face region was divided into several of sub-regions using the detected facial landmark points at different scales and color channels, then these sub-regions were used for training different networks. 
Xie \textit{et al.} proposed the comparator network \cite{ComparatorNet_ECCV2018} that used attention mechanism based on multiple discriminative local sub-regions, and compared local descriptors between pairs of faces. 
Han \textit{et al.} \cite{Contrastive_CNN_ECCV2018} proposed the contrastive convolution which specifically focused on the distinct (contrastive) characteristics between two faces, where it tried to find the differences and put more attention for better discrimination of two faces. For example, the best contrastive feature for distinguishing two images of Stephen Fry and Brad Pitt might be ``crooked nose''.
Kang \textit{et al.} proposed the pairwise relational network (PRN) \cite{PRN_FR_ECCV2018} that made all possible pairs of local appearance features, then each pair of local appearance features is used for capturing relational features. In addition, the PRN was constrained by the face identity state feature embedded from the LSTM-based sub-network to represent face identity. However, these methods largely were dependent on the accuracy of facial landmark detector and it did not use the importance of facial parts.

To overcome these demerits, we propose a novel face recognition method, called Attentional Feature-pair Relation Network (AFRN), which represents the face by the relevant pairs of local appearance block features with their attention scores: 1) the AFRN represents the face by all possible pairs of the 9$\times$9 local appearance block features, 2) the importance of each pair is considered by the attention map that is obtained from the low-rank bilinear pooling, and each pair is weighted by its corresponding attention score, 3) we select top-$K$ pairs of local appearance block features as relevant facial information and drop the remaining irrelevant, 4) The weighted top-$K$ pairs are propagated to extract the joint feature-pair relation by using bilinear attention network.
\figurename{~\ref{fig:fig1}} shows the working principle of the proposed AFRN.

The main contributions of this paper can be summarized as follows:\vspace{-0.2cm}
\begin{itemize}[leftmargin=+0.2in]
	\item \textbf{Landmark free local appearance representation}: we propose a novel face recognition method using the attentional feature-pair relation network (AFRN) which represents the face by the relevant pairs of local appearance block features with their attention scores to captures the unique and discriminative feature-pair relations to classify face images among different identities.\vspace{-0.2cm}
	\item \textbf{Importance of pairs and removing irrelevant pairs}: to consider the importance of each pair, we compute the bilinear attention map by using the low-rank bilinear pooling, and each pair is weighted by its attention score, then we select top-$K$ pairs of local appearance block features as relevant facial information and drop the remaining irrelevant. The weighted top-$K$ pairs are propagated to extract the joint relational feature by using bilinear attention network.\vspace{-0.2cm}
	\item We show that the proposed AFRN improves effectively the accuracy of both face verification and face identification.\vspace{-0.2cm}
	\item To investigate the effectiveness of the AFRN, we present extensive experiments on the public available datasets such as LFW \cite{LFW}, YTF \cite{YTF_CVPR2011}, Cross-Age LFW (CALFW), Cross-Pose LFW (CPLFW), Celebrities in Frontal-Profile in the Wild (CFP) \cite{CFP-FP}, AgeDB \cite{AgeDB}, IARPA Janus Benchmark-A (IJB-A) \cite{IJB-A_CVPR2015}, IARPA Janus Benchmark-B (IJB-B) \cite{IJB-B_CVPRW2017}, and IARPA Janus Benchmark-C (IJB-C) \cite{IJB-C}. % CALFW/CPLFW, CFP, AgeDB... IJB-C 추가
\end{itemize}

\section{Proposed Methods}\label{sec:AFRN}
In this section, we describe the proposed methods in detail including a facial feature encoding network, attentional feature-pair relation network, top-\textit{K} pairs selection and attention allocation.

\subsection{Facial Feature Encoding Network}\label{sec:facial_feature_encoding_network}
A facial feature encoding network is a backbone neural network which encodes a face image into deeply embedded features. We employ the ResNet-101 network \cite{ResNet} and modify it due to the differences of input resolutions, the size of convolution filters, and the size of output feature maps. A detailed architecture configuration of the modified ResNet-101 is summarized in \tablename{~\ref{tab:table_1}}. The non-linear activation outputs of the last convolution layer (\textit{conv5\_3}) are used as the feature maps of facial appearance representation. 

\begin{table}[t]
	\centering
	\caption{The detailed configuration of the modified ResNet-101 for the facial feature encoding network.}
	\label{tab:table_1}
	\resizebox{\linewidth}{!} {
		\begin{tabular}{@{}c|c|c@{}} 
			\toprule
			\textbf{Layer name} & \textbf{Output size} & \textbf{Filter (kernel, \#, stride)} \\ \hline
			\textit{conv1} & $140\times140$ & $5\times5$, 64, 1 \\ 
			\textit{pool} & $70\times70$ & $3\times3$ max pool, -, 2 \\ 
			\textit{conv2\_x} & $70\times70$ & $[(1\times1, 64), (3\times3, 64), (1\times1, 256)] \times 3$ \\ 
			\textit{conv3\_x} & $35\times35$ & $[(1\times1, 128), (3\times3, 128), (1\times1, 512)] \times 4$ \\ 
			\textit{conv4\_x} & $18\times18$ & $[(1\times1, 256), (3\times3, 256), (1\times1, 1024)] \times 23$ \\ 
			\textit{conv5\_x} & $9\times9$ & $[(1\times1, 512), (3\times3, 512), (1\times1, 2048)] \times 3$ \\ \bottomrule
		\end{tabular} 
	}
\end{table}

\subsection{Facial Local Feature Representation}\label{sec:facial_local_feature_representation}
\begin{figure}[t]	
	\centering	
	\subfigure[]{
		\includegraphics[width=.3\linewidth]{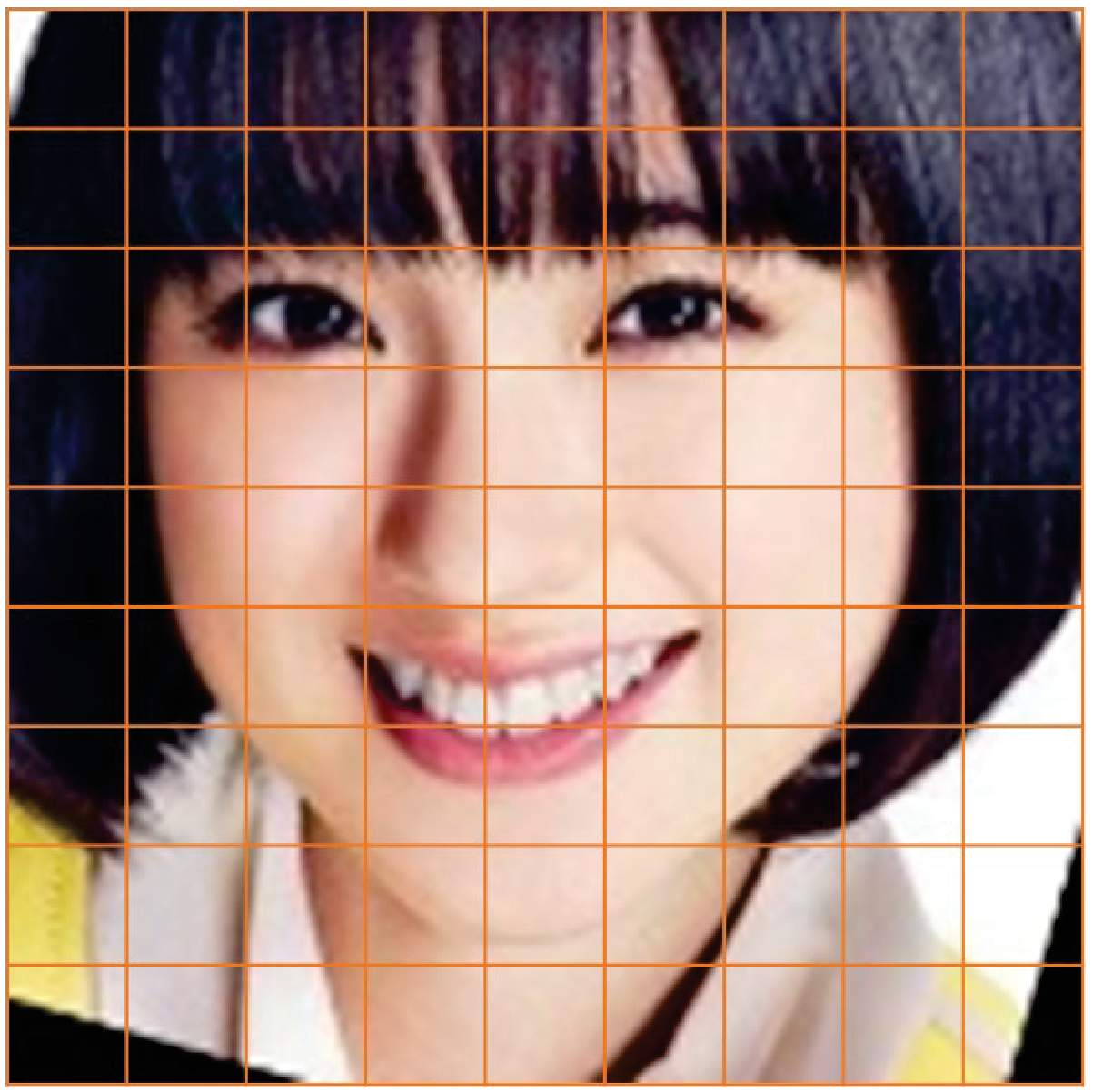}
		\label{fig:fig2_a}
	}
	\subfigure[]{
		\includegraphics[width=.3\linewidth]{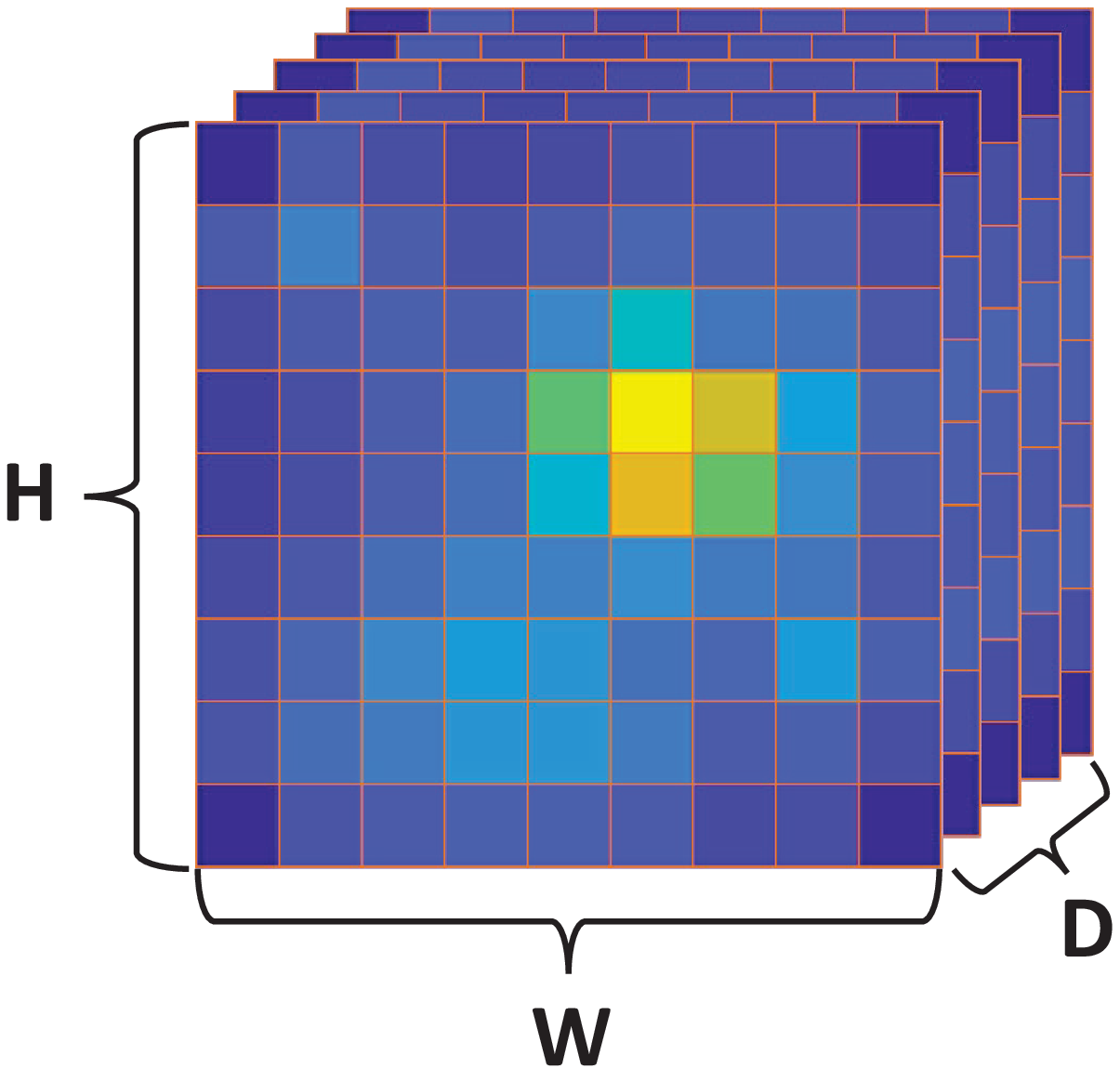}
		\label{fig:fib2_b}
	}
	\caption{Facial local blocks: (a) input face image. (b) facial local blocks on the feature maps.}
	\label{fig:fig2}
\end{figure}

The activation outputs of the convolution layer can be formulated as a tensor of the size $H\times W\times D$, where $H$ and $W$ denote the height and width of each feature map, and $D$ denotes the number of channels in feature maps. 
Essentially, the convolution layer divides the input image into $H\times W$ sub-regions and uses $D$-dimensional feature maps to describe the facial part information within each sub-region. 
For clarity, since the activation outputs of the convolutional layer can be viewed as a 2-D array of $D$-dimensional features, we use each $D$-dimensional local appearance block feature $\boldsymbol{f}_{i}$ of the $H\times W$ sub-regions as the local feature representation of the $i$-th facial part. %we correspond each of the $H\times W$ sub-regions to a facial part, and correspond the $D$-dimensional local appearance block feature $\boldsymbol{f}_{i}$ to the $i$-th facial part. 
Based on the feature map in the \textit{conv5\_3} residual block, the face region is divided into $81$ local blocks ($9\times9$ resolution) (\figurename{~\ref{fig:fig2}}), where each local block is used for the local appearance block feature of a facial part. Therefore, we extract totally 81 local appearance block features $\boldsymbol{A} = \left\{\boldsymbol{f}_{i}| i=1, \cdots, 81\right\}$, where $\boldsymbol{f}_{i} \in \mathbb{R}^{2,048}$ in this work.

\subsection{Attentional Feature-Pair Relation Network}\label{sec:attentional_feature-pair_relation_network}
\begin{figure}[t]
	\centering
	\includegraphics[width=\linewidth]{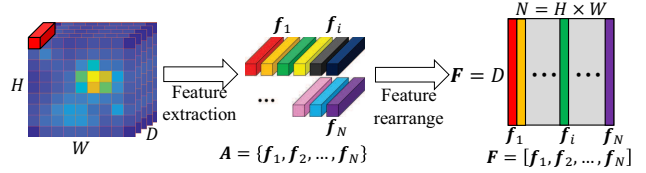}
	\caption{Facial feature rearrangement.}
	\label{fig:fig3}
\end{figure}

The attentional feature-pair relation network (AFRN) is based on the low-rank bilinear pooling \cite{Unitary_Attention} which provides richer representations than linear models and finds attention distributions by considering every pair of features. 
The AFRN aims to represent a separable and discriminative feature-pair relation which is pooled by feature-pair attention scores of feature-pair relations among all possible pairs of given local appearance block features. 
Thus, the AFRN exploits attentional feature-pair relations between all pairs of local appearance block features while extracts a joint feature-pair relation for pairs of local appearance block features. 
\\
\\
\noindent\textbf{Rearrange Local Appearance Block Features.}
To obtain a feature-pair bilinear attention map and a joint feature-pair relation for all of pairs of local appearance block features, we first rearrange a set of local appearance block features $\boldsymbol{A}$ into a matrix form $\boldsymbol{F}$ by stacking each local appearance block feature $\boldsymbol{f}_{i}$ in column direction, $\boldsymbol{F} = [\boldsymbol{f}_{1}, \cdots, \boldsymbol{f}_{i}, \cdots, \boldsymbol{f}_{N}] \in \mathbb{R}^{D\times N}$, where $N$ ($=H\times W$) is the number of local appearance block features (\figurename{~\ref{fig:fig3}}).
\\
\\
\noindent\textbf{Feature-pair Bilinear Attention Map.}
An attention mechanism provides an efficient way to improve accuracy and reduce the number of input features at the same time by selectively utilizing given information. 
We adopt the feature-pair bilinear attention map $\boldsymbol{\mathcal{A}} \in \mathbb{R}^{N\times N}$. To obtain $\boldsymbol{\mathcal{A}}$, we compute a logit of the \textit{softmax} for a pair $\boldsymbol{p}_{i,j}$ between local appearance block features $\boldsymbol{F}_{i}$ and $\boldsymbol{F}_{j}$ as:
\begin{equation}
\boldsymbol{\mathcal{A}}_{i, j} = \boldsymbol{p}^{T}\left(\sigma\left(\boldsymbol{U}^{'T}\boldsymbol{F}_{i}\right)\circ\sigma\left(\boldsymbol{V'}^{T}\boldsymbol{F}_{j}\right)\right),
\label{eq:eq_4}
\end{equation}
where $\boldsymbol{\mathcal{A}}_{i,j}$ is the logit of the \textit{softmax} for $\boldsymbol{p}_{i,j}$ and is the output of low-rank bilinear pooling. $\boldsymbol{U}^{'} \in \mathbb{R}^{D\times L^{'}}$, $\boldsymbol{V}^{'} \in \mathbb{R}^{D\times L^{'}}$, and $\boldsymbol{p} \in \mathbb{R}^{L^{'}}$, where $L^{'}$ is the dimension of the reduced and pooled features by linear mapping $\boldsymbol{U}^{'}$, $\boldsymbol{V}^{'}$ and pooling $\boldsymbol{p}$ in the low-rank bilinear pooling. 
$\sigma$ and $\circ$ denote the ReLU \cite{ReLU} non-linear activation function and Hadamard product (element-wise multiplication), respectively. 
%$\sigma$ denotes the ReLU \cite{ReLU} non-linear activation function, and $\circ$ denotes Hadamard product (element-wise multiplication). 
To obtain $\boldsymbol{\mathcal{A}}$, the \textit{softmax} function is applied element-wisely to each logit $\boldsymbol{\mathcal{A}}_{i,j}$. All above operations can be rewritten as a matrix form: 
\begin{equation}
\boldsymbol{\mathcal{A}} = softmax\left(\left(\left(\boldsymbol{\mathbbm{1}}\cdot\boldsymbol{p}^{T}\right)\circ\sigma\left(\boldsymbol{F}^{T}\boldsymbol{U}^{'}\right)\right)\cdot\sigma\left(\boldsymbol{V}^{'T}\boldsymbol{F}\right)\right),
\label{eq:eq_5}
\end{equation}
where $\boldsymbol{\mathbbm{1}} \in \mathbb{R}^{N}$. \figurename{~\ref{fig:fig5} illustrates a process of the proposed feature-pair bilinear attention map. 
\\	
\begin{figure}[t]
	\centering
	\includegraphics[width=\linewidth]{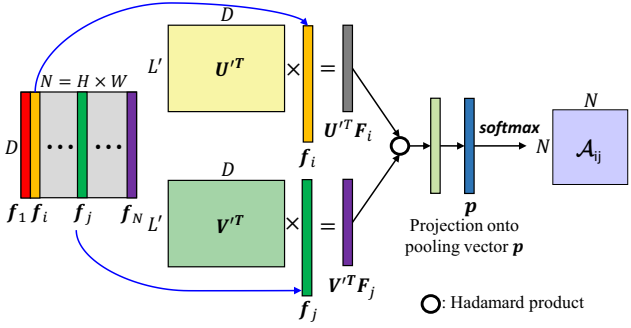}
	\caption{A process of the proposed feature-pair bilinear attention map.}
	\label{fig:fig5}
\end{figure}
\\
\noindent\textbf{Joint Feature-pair Relation.}\label{sec:joint_feature-pair_relation}
To extract a joint feature-pair relation for all of pairs of local appearance block features and reduce the number of pairs of local appearance block features, we use the low-rank bilinear pooling with the feature-pair bilinear attention map $\boldsymbol{\mathcal{A}}$ as:
\begin{equation}
\boldsymbol{r}_{l}^{'}=\sigma\left(\boldsymbol{F}^{T}\boldsymbol{U}\right)_{l}^{T}\cdot\boldsymbol{\mathcal{A}}\cdot\sigma\left(\boldsymbol{F}^{T}\boldsymbol{V}\right)_{l},
\label{eq:eq_1}
\end{equation}
where $\boldsymbol{U}\in\mathbb{R}^{D\times L}$ and $\boldsymbol{V}\in\mathbb{R}^{D\times L}$ are linear mappings. $L$ is the dimension of the reduced and pooled features by pooling and linear mapping matrix $\boldsymbol{U}$ and $\boldsymbol{V}$ in the low-rank bilinear pooling for the feature-pair relation.
$(\boldsymbol{F}^{T}\boldsymbol{U})_{l}\in\mathbb{R}^{N}$, $(\boldsymbol{F}^{T}\boldsymbol{V})_{l}\in\mathbb{R}^{N}$, and $\boldsymbol{r}_{l}^{'}$ denotes the $l$-th element of the intermediate feature-pair relation. The subscript $l$ for the matrices indicates the index of column. $\sigma$ denotes the ReLU \cite{ReLU} non-linear activation function. Eq. \eqref{eq:eq_1} can be viewd as a bilinear model for the pairs of local appearance block features where $\boldsymbol{\mathcal{A}}$ is a bilinear weight matrix (\figurename{~\ref{fig:fig4}}). Therefore, we can rewrite Eq. \eqref{eq:eq_1} as:
\begin{equation}
\boldsymbol{r}_{l}^{'} = \sum_{i=1}^{N}\sum_{j=1}^{N}\boldsymbol{\mathcal{A}}_{i,j}\cdot\sigma\left(\boldsymbol{F}_{i}^{T}\boldsymbol{U}_{l}\right)\cdot\sigma\left(\boldsymbol{V}_{l}^{T}\boldsymbol{F}_{j}\right),
\label{eq:eq_2}
\end{equation}
where $\boldsymbol{F}_{i}$ and $\boldsymbol{F}_{j}$ denote the $i$-th local appearance block feature and the $j$-the local appearance block features of input $\boldsymbol{F}$, respectively. $\boldsymbol{U}_{l}$ and $\boldsymbol{V}_{l}$ denote the $l$-th columns of $\boldsymbol{U}$ and $\boldsymbol{V}$ matrices, respectively. $\boldsymbol{\mathcal{A}}_{i,j}$ denotes an element in the $i$-th row and $j$-th column of $\boldsymbol{\mathcal{A}}$.

Finally, the joint feature-pair relation $\tilde{\boldsymbol{r}}$ is obtained by projection $\boldsymbol{r}^{'}$ onto a learnable pooling matrix $\boldsymbol{P}$:
\begin{equation}
\tilde{\boldsymbol{r}} = \boldsymbol{P}^{T}\boldsymbol{r}{'},
\label{eq:eq_3}
\end{equation}
where $\tilde{\boldsymbol{r}} \in \mathbb{R}^{C}$ and $\boldsymbol{P} \in \mathbb{R}^{L\times C}$.
$C$ is the dimension of the joint feature-pair relation by pooling $\boldsymbol{P}$ to obtain the final joint feature-pair relation $\tilde{\boldsymbol{r}}$.

\begin{figure*}[t]
	\centering
	\includegraphics[width=0.7\textwidth]{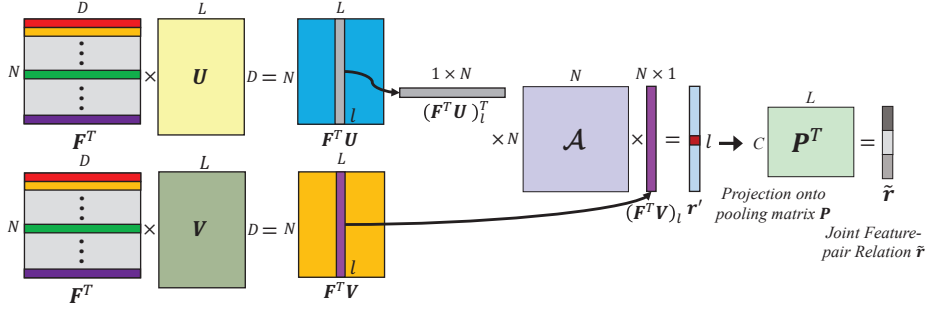}
	\caption{The joint feature-pair relation.}
	\label{fig:fig4}
\end{figure*}

\subsection{Pair Selection and Attention Allocation}\label{sec:pair_selection_attention_allocation}
Only some facial part pairs are relevant to face recognition and irrelevant ones may cause over-fitting of the neural network. We need to select relevant pairs of local appearance block features, therefore we select them with top-$K$ feature-pair bilinear attention scores as: 
\begin{equation}
\boldsymbol{\mathbf{\varPhi}} = \left\{\boldsymbol{p}_{i,j}|\boldsymbol{\mathcal{A}}_{i,j} \text{ ranks top }K\text{ in }\boldsymbol{\mathcal{A}}\right\},
\label{eq:eq_6}
\end{equation}
where $\boldsymbol{p}_{i,j}$ is the selected pair of $\boldsymbol{F}_{i}$ and $\boldsymbol{F}_{j}$ with a top-$K$ feature-pair attention score.

Different pairs of local appearance block features always have equal value scale, yet they offer different contributions on face recognition. So, we should rescale the pairs of local appearance block features to reflect their indeed influence. Mathematically, it is modeled as multiplying the corresponding feature-pair bilinear attention score. Therefore, we can substitute Eq. \eqref{eq:eq_2} as 
\begin{equation}
\boldsymbol{r}_{l}^{'} = \sum_{k=1}^{K}\boldsymbol{\mathcal{A}}_{w_{i}(k), w_{j}(k))}\cdot\sigma\left(\boldsymbol{F}_{w_{i}(k)}^{T}\boldsymbol{U}_{l}\right)\cdot\sigma\left(\boldsymbol{V}_{l}^{T}\boldsymbol{F}_{w_{j}(k)}\right),
\label{eq:eq_7}
\end{equation}
where $w_{i}(k)$ and $w_{j}(k)$ are $i$ and $j$ indexes of the $k$-th pair $\boldsymbol{p}_{i,j}$ in $\boldsymbol{\varPhi}$. $K$ denotes the number of the selected pairs by the pair selection layer.

Because Eq. \eqref{eq:eq_6} is not a differentiable function, it has no parameter to be updated and only conveys gradients from the latter layer to the former layer during back-propagation. The gradients of the selected pairs of local appearance block features will be copied from latter layer to the former layer and the gradients of the dropped pairs of local appearance block features will be discarded by setting the corresponding values to zero.

After the pair selection and attention allocation, the weighted pairs of local appearance block features are propagated the next step to extract the joint feature-pair relation.
The joint feature-pair relation $\tilde{\boldsymbol{r}}$ is fed into two-layered multi-layer perceptron (MLP) $\mathcal{F}_{\boldsymbol{\theta}}$ followed by the loss function. We use the $1,024$ dimensional output vector of the last fully connected layer of $\mathcal{F}_{\boldsymbol{\theta}}$ as a final face representation.

\section{Experiments}\label{sec:experiments}
In this section, we describe the training dataset, validation set, and implementation details. 
We also demonstrate the effectiveness of the proposed AFRN on the LFW \cite{LFW}, YTF \cite{YTF_CVPR2011}, IJB-A \cite{IJB-A_CVPR2015} and IJB-B \cite{IJB-B_CVPRW2017} datasets.

\subsection{Training Dataset}\label{sec:train_data}
We use the VGGFace2 \cite{VGG2Face} dataset which has 3.2M face images from 8,631 unique persons. We detect face regions and their facial landmark points by using the multi-view face detector \cite{FD_YOON2018} and deep alignment networks (DAN) \cite{DAN_CVPRW2017}. When detection is failed, we just discard that images and totally remove 24,160 face images from 6,561 subjects. Then, we have roughly 3.1M face images of 8,630 unique persons as the refined dataset. We divide this dataset into two sets: one for training set having roughly 2.8M face images, and another for validation set with 311,773 face images which are selected randomly about 10\% from each subject.
We use 68 facial landmark points for the face alignment. All of faces in both the training and validation sets are aligned to canonical faces by using the face alignment method in \cite{PRN_FR_ECCV2018}. 
The faces with 140$\times$140 resolutions are used and each pixel is normalized by dividing 255 to be in a range of $[0, 1]$.

\subsection{Implementation Details}\label{sec:implementation}
We extract 81 local appearance block features on the 9$\times$9$\times$2,048 feature maps in \textit{conv5\_3} residual block of the facial feature encoding network, and each local appearance block feature has 2,048 dimensions. 
Thus, the size of local appearance block features is $D=2,048$ and the number of local appearance block features is $N=81$. The size of the rearranged local appearance block features $\boldsymbol{F}$ is $\mathbb{R}^{2,048\times 81}$, the size $C$ of the joint feature-pair relation is 1,024, which is equal to the rank $L$ of the AFRN, and the rank $L^{'}$ of the feature-pair bilinear attention map is also $1,024$. Every linear mapping ($\boldsymbol{U}$, $\boldsymbol{V}$, $\boldsymbol{U}^{'}$, $\boldsymbol{V}^{'}$, and $\boldsymbol{P}$) is regularized by the Weight Normalization \cite{Weight_Normalization_NIPS2016}.
We use the two-layered MLP consisting of $1,024$ units per layer with Batch Normalization (BN) \cite{BatchNormal} and ReLU \cite{ReLU} non-linear activation functions for $\mathcal{F}_{\boldsymbol{\theta}}$.

The proposed AFRN is optimized by jointly using the triplet ratio $L_{t}$, pairwise $L_{p}$, and identity preserving $L_{id}$ loss functions proposed in \cite{PIMNet_CVPRW2017} over the ground-truth identity labels. 
Adamax optimizer \cite{Adamax_ICLR2015}, a variant of Adam based on infinite norm, is used. 
The learning rate is $\min(i\times 10^{-3}, 4\times 10^{-3})$ where $i$ is the number of epochs starting from 1, then after 10 epochs, the learning rate is decayed by $0.25$ for every 2 epochs up to 13 epochs, \textit{i.e.} $1\times 10^{-3}$ for 11-th and $2.5\times 10^{-4}$ for 13-th epoch.
%The learning rate is $\min(ie^{-3}, 4e^{-3})$ where $i$ is the number of epochs starting from 1, then after 10 epochs, the learning rate is decayed by $1/4$ for every 2 epochs up to 13 epochs, \textit{i.e} $1e^{-3}$ for 11-th and $2.5e^{-4}$ for 13-th epoch. 
We clip 2-norm of vectorized gradient to $0.25$.
We achieve the best results by setting the weight factors of loss functions as 1, 0.5, and 1 for $L_{t}$, $L_{p}$, and $L_{id}$ by a grid search, respectively.
We set the mini-batch size as 120 on four NVIDIA Titan X GPUs.

\subsection{Ablation Study}\label{sec:ablation_studies}
We conduct several experiments to analyze the proposed AFRN on the LFW \cite{LFW} and YTF \cite{YTF_CVPR2011} datasets. Following the test protocol of \textit{unrestricted with labeled outside data} \cite{LFWTechUpdate}, we test the proposed AFRN on the LFW and YTF by using a squared $L_{2}$ distance threshold to determine the classification of \textit{same} and \textit{different}, and report the results in \tablename{~\ref{tab:table_2} and \ref{tab:table_3}, and then discuss results in detail.

\noindent\textbf{Effects of Feature-pair Selection.}
In the feature-pair selection layer, we need to decide top-$K$ local appearance pairs that we propagate to the next step. We perform an experiment to evaluate the effect of $K$. We train the AFRN model on the refined VGGFace2 training set with different value of $K$. The accuracy on validation set is reported in \figurename{~\ref{fig:fig7}}. When $K$ increases, the accuracy of our AFRN model increased until $K=442$ (97.4\%). After that, the accuracy of our model starts to drop. When $K$ equals to 1,200, it is equivalent to not using the feature-pair selection layer in a face region. The performance in this case is 2.3\% %6.4\% 
lower than the highest accuracy. This implies that it is important to reject irrelevant the pairs of local appearance block features.

\begin{figure}[t]
	\centering
	\includegraphics[width=0.9\linewidth]{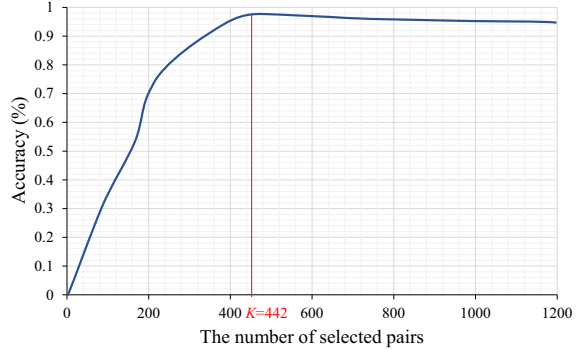}
	\caption{Accuracy plot with the different number $K$ of feature-pair on the validation set.}
	\label{fig:fig7}
\end{figure}
%\vspace{-0.3cm}
%\paragraph{Effects of Feature-pair Bilinear Attention.}

\noindent\textbf{Effects of Feature-pair Bilinear Attention.}
To evaluate the effects of the feature-pair bilinear attention in the proposed AFRN, we perform several experiments on the validation set, LFW and YTF datasets. We consider the attentional feature-pair relation network without the feature-pair selection layer, which means that we use all pairs of local appearance block features for face recognition. We achieve 95.1\% accuracy on the validation set, 99.71\% accuracy on the LFW, and 96.1\% accuracy on the YTF, respectively (\tablename{~\ref{tab:table_2} (b)} and \figurename{~\ref{fig:fig8}}). 
We use the normalized face image which include the background regions and is not cropped a face region tightly (see \figurename{~\ref{fig:fig2}}). When not using pair selection, we observe that attention scores for pairs between background regions and face regions are not zero, and the accuracy is degraded in comparison with the baseline (\tablename{~\ref{tab:table_2} (a) and \figurename{~\ref{fig:fig8}}). 
It indicates that all possible pairs are not necessarily for face recognition. Therefore, we need to remove irrelevant pairs of local appearance block features.

Then, we consider the attentional feature-pair relation network with the feature-pair selection layer of $K=442$.  
We achieve 97.4\% accuracy on the validation set, 99.85\% accuracy on the LFW, and 97.1\% accuracy on the YTF, respectively (\tablename{~\ref{tab:table_2} (c)} and \figurename{~\ref{fig:fig8}}). 
The experimental results show that the AFRN with top-\textit{K} selection layer outperforms the current state-of-the-art accuracies as 99.78\% (ArcFace \cite{ArcFace}) on the LFW dataset and 96.3\% (PRN \cite{PRN_FR_ECCV2018}) on the YTF dataset.

\begin{table}[t]
	\centering
	\caption{Effects of the feature-pair selection by the feature-pair bilinear attention on the validation set, LFW and YTF dataset.}
	\label{tab:table_2}
	\resizebox{0.9\linewidth}{!} {
		\begin{tabular}{@{}l|c|c|c@{}}
			\toprule
			\multicolumn{1}{c|}{Method}  & Val. set & LFW & YTF \\ \midrule
			(a) Baseline & 94.2 &  99.60 & 95.1 \\ \midrule
			(b) Feature-pair Attention w/o Pair Selection &  95.1 &  99.71 & 96.1 \\
			(c) Feature-pair Attention w/ Pair Selection  &  \textbf{97.4} &  \textbf{99.85} & \textbf{97.1} \\ \midrule
			(d) ArcFace \cite{ArcFace} &  - &  99.78 & - \\
			(e) PRN \cite{PRN_FR_ECCV2018} &  - &  99.76 & 96.3 \\ \bottomrule		
		\end{tabular}
	}
\end{table}
\begin{figure}[t]
	\centering
	\includegraphics[width=0.8\linewidth]{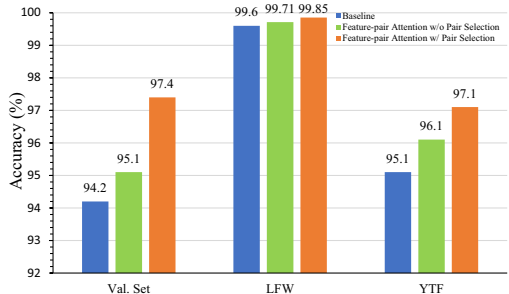}
	\caption{Effects of the feature-pair selection by the feature-pair bilinear attention on the validation set, LFW and YTF datasets.}
	\label{fig:fig8}
\end{figure}
%\vspace{-0.59cm}
%\paragraph{Comparison with Other Attention Mechanisms.}

\noindent\textbf{Comparison with Other Attention Mechanisms.}
To compare with other attention mechanisms, we conduct ablation study with top-\textit{K} pair selection ($K=442$) for comparison with other attention mechanisms including the unitary attention \cite{Unitary_Attention} and co-attention \cite{Co-attention} on the validation set, LFW, and YTF datasets. 
We achieve 97.4\% accuracy on the validation set, 99.85\% accuracy on the LFW, and 97.1\% accuracy on the YTF, respectively (\tablename{~\ref{tab:table_3}). It indicates that the proposed feature-pair bilinear attention shows better accuracy than the other attention mechanisms.
	
\begin{table}[t]
	\centering
	\caption{Comparison results with other attention mechanisms.}
	\label{tab:table_3}
	\resizebox{0.8\linewidth}{!} {
		\begin{tabular}{@{}l|c|c|c@{}}
			\toprule
			\multicolumn{1}{c|}{Method}  & Val. set & LFW & YTF \\ \midrule
			(a) Unitary Attention \cite{Unitary_Attention}&  95.3 &  99.53 & 95.3 \\
			(b) Co-attention \cite{Co-attention} &  96.1 &  99.63 & 95.8 \\
			(c) Feature-pair Bilinear Attention  &  \textbf{97.4} &  \textbf{99.85} & \textbf{97.1} \\
			\bottomrule		
		\end{tabular}
	}
\end{table}

\subsection{Comparison with the State-of-the-art Methods}\label{sec:comparison_SOTA}
%\paragraph{Detailed Settings in the Models.}
\noindent\textbf{Detailed Settings in the Models.}
For fair comparison in terms of the effects of each network module, we train three kinds of models (\textbf{model A}, \textbf{model B}, and \textbf{model C}) using the triplet ratio, pairwise, and identity preserving loss functions \cite{PIMNet_CVPRW2017} jointly over the ground-truth identity labels:
\textbf{model A} is the facial feature encoding network model with only the global appearance feature (\tablename{~\ref{tab:table_1}}).
\textbf{model B} is the AFRN model without the feature-pair selection layer.
\textbf{model C} is the AFRN model with the feature-pair selection layer.
All of convolution layers and fully connected layers used BN and ReLU as non-linear activation functions.

\noindent\textbf{Experiments on the IJB-A dataset.}
We evaluate the proposed models on the IJB-A dataset \cite{IJB-A_CVPR2015} which contains face images and videos captured from the unconstrained environments. The IJB-A dataset is very challenging due to its full pose variation and wide variations in imaging conditions, and contains 500 subjects with 5,397 images and 2,042 videos in total, and 11.4 images and 4.2 videos per subject on average. We detect the face regions using the face detector \cite{FD_YOON2018} and the facial landmark points using DAN \cite{DAN_CVPRW2017} landmark point detector, and then aligned the face image by using the alignment method in \cite{PRN_FR_ECCV2018}.

\begin{table*}[t]
	\caption{Comparison of performances of the proposed AFRN method with the \textit{state-of-the-art} on the IJB-A dataset. For verification, TAR \textsl{vs.} FAR are reported. For identification, TPIR \textsl{vs.} FPIR and the Rank-N accuracies are presented.}\label{tab:table_4}
	\resizebox{\textwidth}{!} {
		\begin{tabular}{@{}llllllllll@{}}
			\toprule
			\multicolumn{1}{c}{\multirow{2}{*}{Method}}            & \multicolumn{3}{c}{1:1 Verification TAR} & \multirow{2}{*}{} & \multicolumn{5}{c}{1:N Identification TPIR}  \\ \cmidrule(lr){2-4} \cmidrule(l){6-10}
			\multicolumn{1}{c}{}                                   & FAR=0.001        &  FAR=0.01         &  FAR=0.1          &  &  FPIR=0.01        &  FPIR=0.1         &  Rank-1           & Rank-5           &  Rank-10           \\ \midrule
			Pose-Aware Models \cite{POSE_AWARE_CVPR2016}           & $0.652\pm0.037$  &  $0.826\pm0.018$  &  -                &  &  -                &  -                &  $0.840\pm0.012$  &  $0.925\pm0.008$  &  $0.946\pm0.005$  \\
			All-in-One \cite{All-in-one_CNN_FR}                    & $0.823\pm0.02$   &  $0.922\pm0.01$   &  $0.976\pm0.004$  &  &  $0.792\pm0.02$   &  $0.887\pm0.014$  &  $0.947\pm0.008$  &  $0.988\pm0.003$  &  $0.986\pm0.003$  \\
			NAN \cite{NAN_CVPR2017}                                & $0.881\pm0.011$  &  $0.941\pm0.008$  &  $0.978\pm0.003$  &  &  $0.817\pm0.041$  &  $0.917\pm0.009$  &  $0.958\pm0.005$  &  $0.980\pm0.005$  &  $0.986\pm0.003$  \\			
			VGGFace2 \cite{VGG2Face}                               & $0.904\pm0.020$  &  $0.958\pm0.004$  &  $0.985\pm0.002$  &  &  $0.847\pm0.051$  &  $0.930\pm0.007$  &  $0.981\pm0.003$  &  $0.994\pm0.002$  &  $0.996\pm0.001$  \\
			VGGFace2\_ft \cite{VGG2Face}                	       & $0.921\pm0.014$  &  $0.968\pm0.006$  &  $0.990\pm0.002$  &  &  $0.883\pm0.038$  &  $0.946\pm0.004$  &  $0.982\pm0.004$  &  $0.993\pm0.002$  &  $0.994\pm0.001$  \\
			PRN \cite{PRN_FR_ECCV2018}       					   & $0.901\pm0.014$  &  $0.950\pm0.006$  &  $0.985\pm0.002$  &  &  $0.861\pm0.038$  &  $0.931\pm0.004$  &  $0.976\pm0.003$  &  $0.992\pm0.003$  &  $0.994\pm0.003$  \\
			PRN${}^{+}$ \cite{PRN_FR_ECCV2018}    				   & $0.919\pm0.013$  &  $0.965\pm0.004$  &  $0.988\pm0.002$  &  &  $0.882\pm0.038$  &  $0.941\pm0.004$  &  $0.982\pm0.004$  &  $0.992\pm0.002$  &  $0.995\pm0.001$  \\
			DR-GAN \cite{DR-GAN_CVPR2017}    				       & $0.539\pm0.043$  &  $0.774\pm0.027$  &  -                &  &  -                &  -                &  $0.855\pm0.015$  &  $0.947\pm0.011$  &  -  \\
			DREAM \cite{DREAM_CVPR2018}    				           & $0.868\pm0.015$  &  $0.944\pm0.009$  &  -                &  &  -                &  -                &  $0.946\pm0.011$  &  $0.968\pm0.010$  &  -  \\
			DA-GAN \cite{DA-GAN_TPAMI2018}    				       & $0.930\pm0.005$  &  $0.976\pm0.007$  &  $0.991\pm0.003$  &  &  $0.890\pm0.039$  &  $0.949\pm0.009$  &  $0.971\pm0.007$  &  $0.989\pm0.003$  &  -  \\ \midrule
			\textbf{model A} (baseline)                            & $\mathbf{0.895\pm0.015}$  &  $\mathbf{0.949\pm0.008}$  &  $\mathbf{0.980\pm0.005}$  &  &  $\mathbf{0.843\pm0.035}$  &  $\mathbf{0.923\pm0.005}$  &  $\mathbf{0.975\pm0.005}$  &  $\mathbf{0.992\pm0.004}$  &  $\mathbf{0.993\pm0.001}$ \\
			\textbf{model B} (AFRN w/o pair selection)        	   & $\mathbf{0.904\pm0.013}$  &  $\mathbf{0.953\pm0.006}$  &  $\mathbf{0.985\pm0.002}$  &  &  $\mathbf{0.869\pm0.038}$  &  $\mathbf{0.935\pm0.004}$  &  $\mathbf{0.981\pm0.003}$  &  $\mathbf{0.993\pm0.003}$  &  $\mathbf{0.994\pm0.002}$ \\
			\textbf{model C} (AFRN w/ pair selection)     		   & $\mathbf{0.949\pm0.013}$  &  $\mathbf{0.985\pm0.004}$  &  $\mathbf{0.998\pm0.002}$  &  &  $\mathbf{0.942\pm0.038}$  &  $\mathbf{0.968\pm0.004}$  &  $\mathbf{0.993\pm0.004}$  &  $\mathbf{0.995\pm0.001}$  &  $\mathbf{0.996\pm0.001}$ \\ \bottomrule
		\end{tabular}
	}
\end{table*}

\begin{figure*}[t]
	\centering
	\subfigure[ROC]{
		\includegraphics[width=0.4\linewidth]{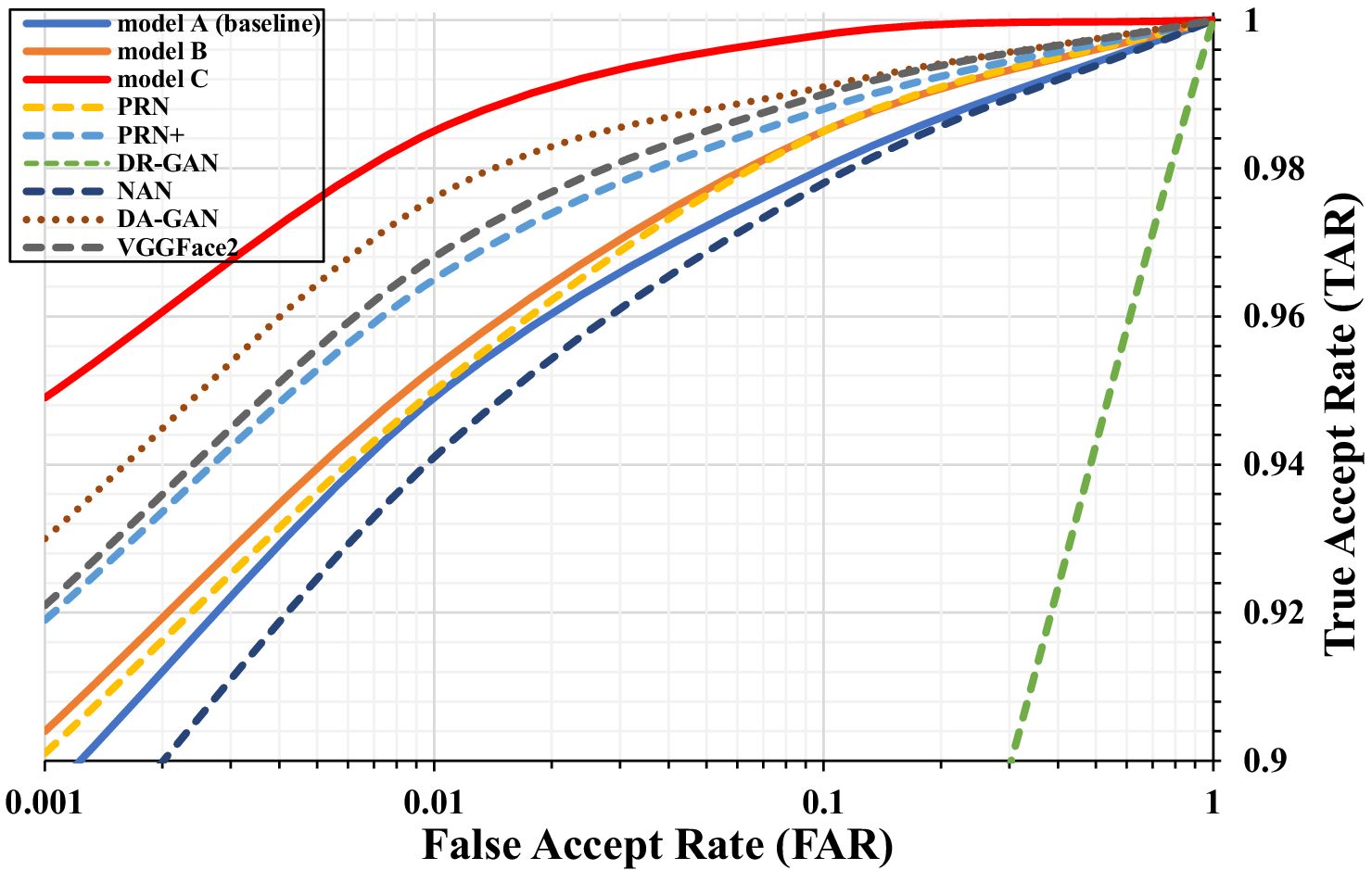}}
	\subfigure[DET]{
		\includegraphics[width=0.4\linewidth]{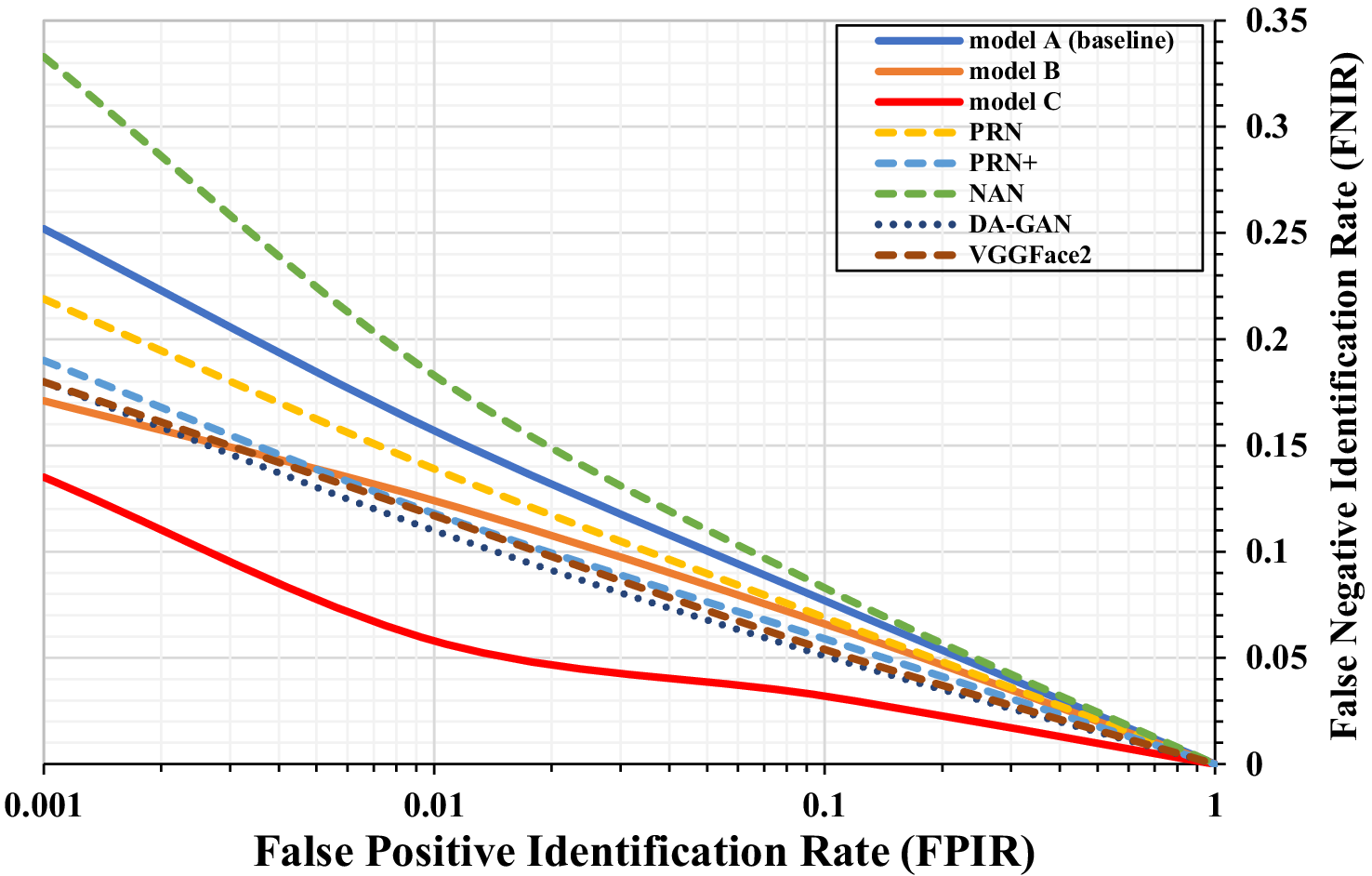}}
	\caption{Comparison of three AFRN models with the \textit{state-of-the-art} methods on the IJB-A dataset (average over 10 splits): (a) ROC (higher is better) and (b) DET (lower is better).}
	\label{fig:fig9}
\end{figure*}

Three models (\textbf{model A}, \textbf{model B}, and \textbf{model C}) are trained on the roughly 2.8M refined VGGFace2 training set, with no people overlapping with subjects in the IJB-A dataset. The IJB-A dataset provides 10 split evaluations with two protocols (1:1 face verification and 1:N face identification). For 1:1 face verification, we report the test results by using true accept rate (TAR) \textit{vs.} false accept rate (FAR) (i.e. receiver operating characteristics (ROC) curve) (\tablename{~\ref{tab:table_4}} and \figurename{~\ref{fig:fig9}} (a)). For 1:N face identification, we report the results by using the true positive identification rate (TPIR) \textit{vs.} false positive identification rate (FPIR) (equivalent to a decision error trade-off (DET) curve) and Rank-N (\tablename{~\ref{tab:table_4}} and \figurename{~\ref{fig:fig9}} (b)). We average all the $1,024$ dimensional output vectors of the last fully connected layer of $\mathcal{F}_{\boldsymbol{\theta}}$ for a media in the template, then we average these media-averaged features to get the final template feature as face representation. All performance evaluations are based on the squared $L_{2}$ distance threshold. 

From the experimental results (\tablename{~\ref{tab:table_4}} and \figurename{~\ref{fig:fig9}}), we have the following observations.
First, compared to \textbf{model A}, \textbf{model B} achieves a consistently superior accuracies (TAR and TPIR) by 0.4-0.9\% for TAR at FAR=0.001-0.1 in verification task, 1.2-2.6\% for TPIR at FPIR=0.01 and 0.1 in identification open set task, and 0.6\% for Rank-1 in identification close set task.
Second, \textbf{model C} shows a consistently higher accuracy than \textbf{model A} by the improvement of 1.8-5.4\% TAR at FAR = 0.001-0.1 in the verification task, 4.5-9.9\% TPIR at FPIR = 0.01-0.1 in the identification open set task, and 1.8\% Rank-1 in the identification close set task.
Third, \textbf{model C} shows a consistently higher accuracy than \textbf{model B} by the improvement of 1.3-4.5\% TAR at FAR = 0.001-0.1 in the verification task, 3.3-7.3\% TPIR at FPIR = 0.01-0.1 in the identification open set task, and 1.5\% for rank-1 in the identification close set task.
Last, although \textbf{model C} is trained from scratch, it outperformed the state-of-the-art method (DA-GAN \cite{DA-GAN_TPAMI2018}) by 0.7-1.9\% TAR at FAR = 0.001-0.1 in the verification task, 2.2\% for Rank-1 on identification close set task, and 5.2\% for TPIR at FPIR = 0.01 in identification open set task on the IJB-A dataset. This validates the effectiveness of the proposed AFRN with the pair selection on the large-scale and challenging unconstrained face recognition.

\begin{table*}[t]
	\caption{Comparison of performances of the proposed AFRN method with the \textit{state-of-the-art} on the IJB-B dataset. For verification, TAR \textsl{vs.} FAR are reported. For identification, TPIR \textsl{vs.} FPIR and the Rank-N accuracies are presented.}
	\label{tab:table_5}
	\resizebox{\textwidth}{!} {
		\begin{tabular}{@{}lllllllllll@{}}
			\toprule
			\multicolumn{1}{c}{\multirow{2}{*}{Method}} & \multicolumn{4}{c}{1:1 Verification TAR} & \multirow{2}{*}{} & \multicolumn{5}{c}{1:N Identification TPIR}  \\ \cmidrule(lr){2-5} \cmidrule(l){7-11}
			\multicolumn{1}{c}{}                        	& FAR=0.00001  &  FAR=0.0001  &  FAR=0.001  &  FAR=0.01  &  &  FPIR=0.01  &  FPIR=0.1  &  Rank-1  &  Rank-5  &  Rank-10  \\ \midrule
			VGGFace2 \cite{VGG2Face}                    	& $0.671$  & $0.800$  & $0.888$  & $0.949$  &  &  $0.706\pm0.047$  &  $0.839\pm0.035$  &  $0.901\pm0.030$  &  $0.945\pm0.016$  &  $0.958\pm0.010$  \\
			VGGFace2\_ft \cite{VGG2Face}                	& $0.705$  & $0.831$  & $0.908$    & $0.956$  &  &  $0.743\pm0.037$  &  $0.863\pm0.032$  &  $0.902\pm0.036$  &  $0.946\pm0.022$  &  $0.959\pm0.015$  \\
			FPN \cite{FPN_Align_ICCVW2017}              	& -  	   & $0.832$  & $0.916$    & $0.965$  &  &  -  				 &  -  				 &  $0.911$  		 &  $0.953$  		 &  $0.975$  		\\
			Comparator Net \cite{ComparatorNet_ECCV2018}    & -        & $0.849$  & $0.937$    & $0.975$  &  &  -  &  -  &  -  &  -  &  -  \\
			PRN \cite{PRN_FR_ECCV2018}        				& $0.692$  & $0.829$  & $0.910$    & $0.956$  &  &  $0.773\pm0.018$  &  $0.865\pm0.018$  &  $0.913\pm0.022$  &  $0.954\pm0.010$  &  $0.965\pm0.013$  \\
			PRN${}^{+}$ \cite{PRN_FR_ECCV2018}    			& $0.721$  & $0.845$  & $0.923$    & $0.965$  &  &  $0.814\pm0.017$  &  $0.907\pm0.013$  &  $0.935\pm0.015$  &  $0.965\pm0.017$  &  $0.975\pm0.007$  \\ \midrule
			\textbf{model A} (baseline)                     & $\mathbf{0.673}$  & $\mathbf{0.812}$  &  $\mathbf{0.892}$  &  $\mathbf{0.953}$  &  &  $\mathbf{0.743\pm0.019}$  &  $\mathbf{0.851\pm0.017}$  &  $\mathbf{0.911\pm0.017}$  &  $\mathbf{0.950\pm0.013}$  &  $\mathbf{0.961\pm0.010}$  \\
			\textbf{model B} (AFRN w/o pair selection)      & $\mathbf{0.706}$  & $\mathbf{0.839}$  &  $\mathbf{0.933}$  &  $\mathbf{0.966}$  &  &  $\mathbf{0.803\pm0.018}$  &  $\mathbf{0.885\pm0.018}$  &  $\mathbf{0.923\pm0.022}$  &  $\mathbf{0.962\pm0.010}$  &  $\mathbf{0.974\pm0.007}$  \\
			\textbf{model C} (AFRN w/ pair selection)      	& $\mathbf{0.771}$  & $\mathbf{0.885}$  &  $\mathbf{0.949}$  &  $\mathbf{0.979}$  &  &  $\mathbf{0.864\pm0.017}$  &  $\mathbf{0.937\pm0.013}$  &  $\mathbf{0.973\pm0.015}$  &  $\mathbf{0.976\pm0.017}$  &  $\mathbf{0.977\pm0.007}$  \\ \bottomrule
		\end{tabular}
	}
\end{table*}
\begin{figure*}[t]
	\centering
	\subfigure[ROC]{
		\includegraphics[width=0.4\linewidth]{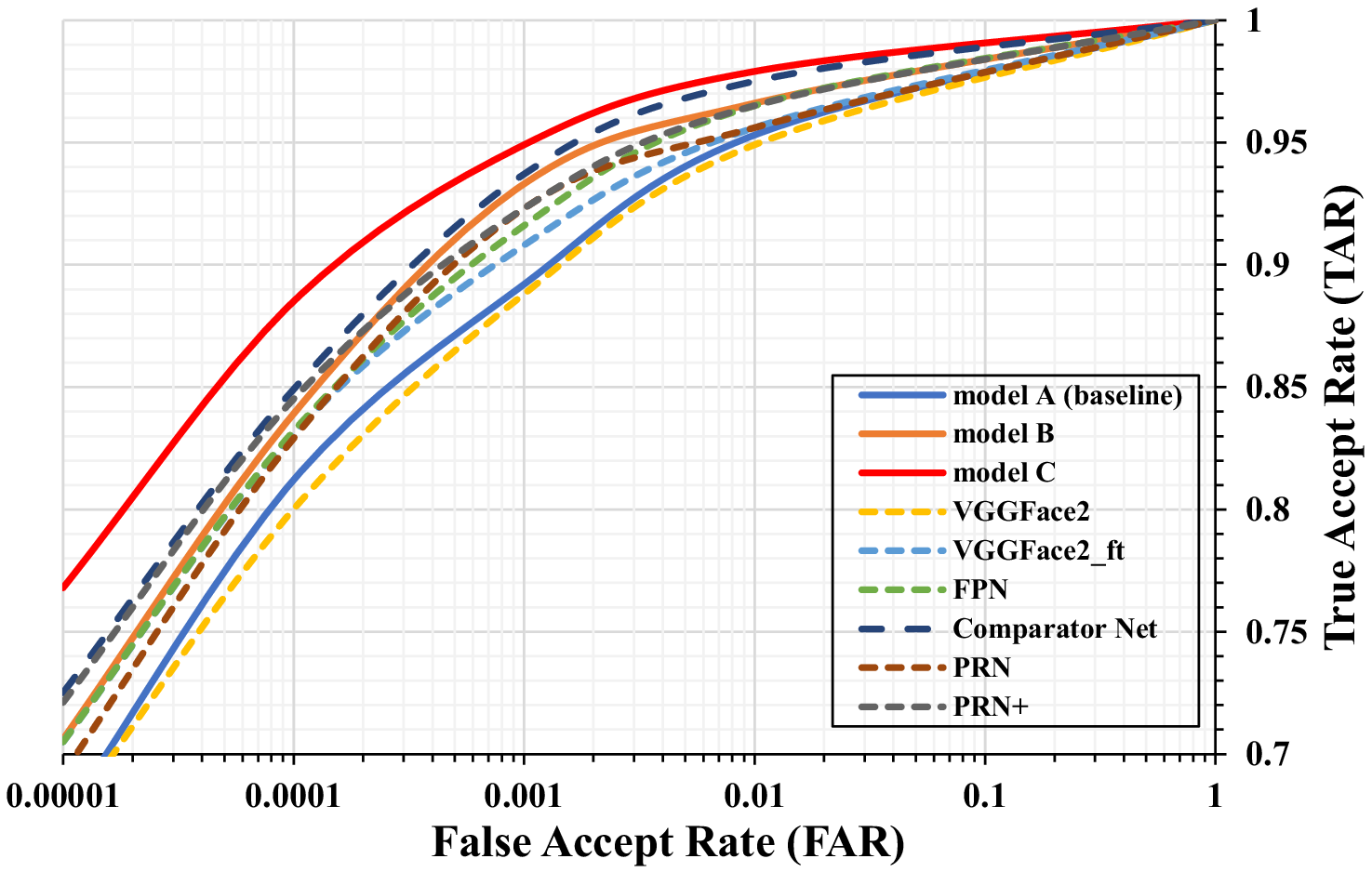}}
	\subfigure[DET]{
		\includegraphics[width=0.4\linewidth]{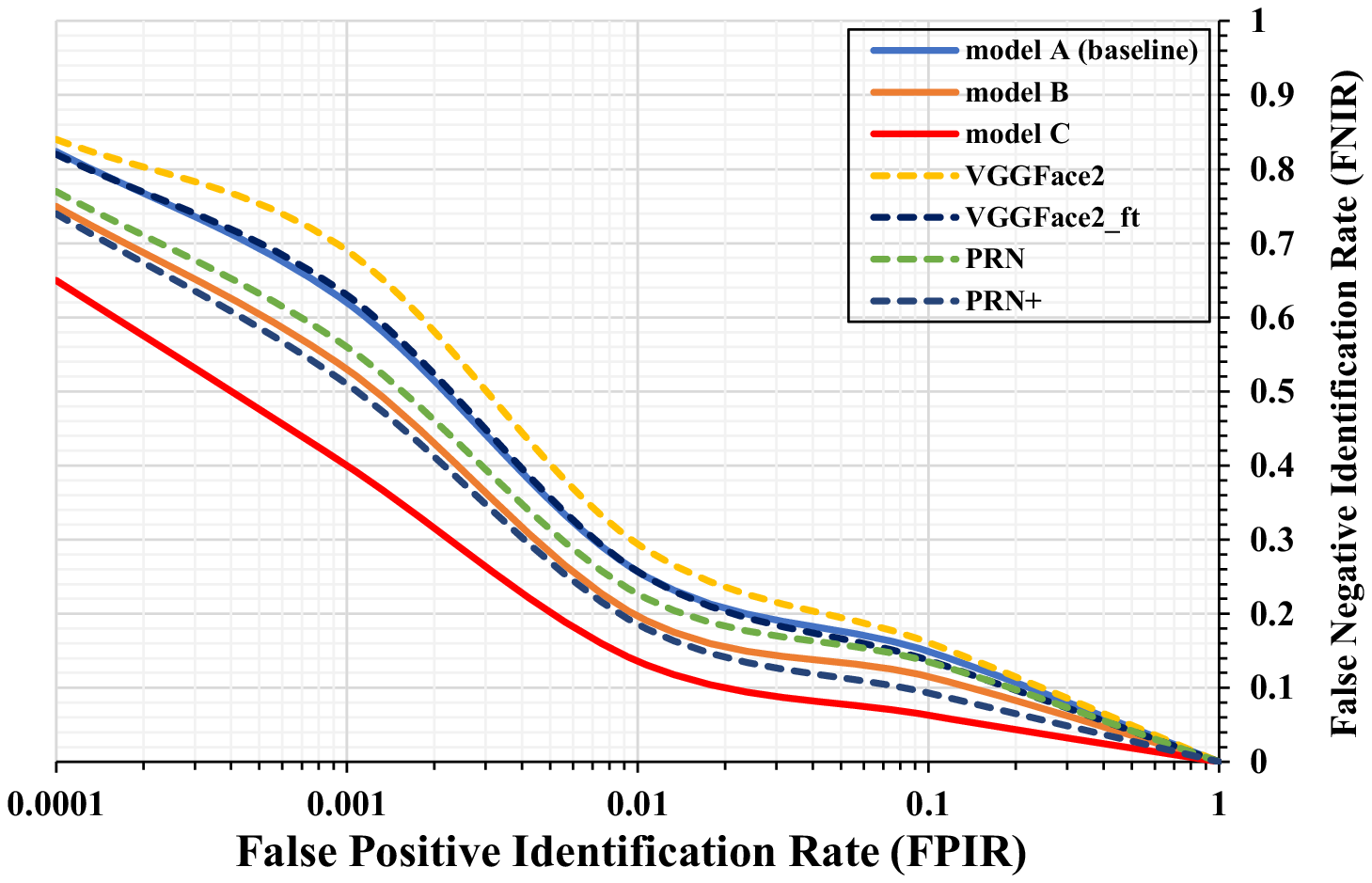}}
	\caption{Comparison of three AFRN models with the \textit{state-of-the-art} methods on the IJB-B dataset: (a) ROC (higher is better) and (b) DET (lower is better).}
	\label{fig:fig10}
\end{figure*}

%\paragraph{Experiments on the IJB-B dataset.}
\noindent\textbf{Experiments on the IJB-B dataset.}
We evaluate the proposed models on the IJB-B dataset \cite{IJB-B_CVPRW2017} which contains face images and videos captured from the unconstrained environments. 
The IJB-B dataset is an extension of the IJB-A dataset, which contains 1,845 subjects with 21.8K still images (including 11,754 face and 10,044 non-face) and 55K frames from 7,011 videos, an average of 41 images per subject.
Because images are labeled with ground truth bounding boxes, we only detect facial landmark points using DAN \cite{DAN_CVPRW2017}, and then aligned face images by using the face alignment method explained in \cite{PRN_FR_ECCV2018}. 

Three models (\textbf{model A}, \textbf{model B}, and \textbf{modelC}) are trained on the roughly 2.8M refined VGGFace2 dataset, with no people overlapping with subjects in the IJB-B dataset.
Unlike the IJB-A dataset, it does not contain any training splits. In particular, we use the 1:1 baseline verification protocol and 1:N mixed media identification protocol for the IJB-B dataset. For 1:1 face verification, we report the test results by using TAR \textit{vs.} FAR (i.e. a ROC curve) (\tablename{~\ref{tab:table_5}} and \figurename{~\ref{fig:fig10}} (a)). For 1:N face identification, we report the results by using TPIR \textit{vs.} FPIR (equivalent to a DET curve) and Rank-N (\tablename{~\ref{tab:table_5}} and \figurename{~\ref{fig:fig10}} (b)). 
We compare three proposed models with VGGFace2 \cite{VGG2Face}, FacePoseNet (FPN) \cite{FPN_Align_ICCVW2017}, Comparator Net \cite{ComparatorNet_ECCV2018}, and PRN \cite{PRN_FR_ECCV2018}.
Similarity to evaluation on the IJB-A, all performance evaluations are based on the squared $L_{2}$ distance threshold.

From the experimental results (\tablename{~\ref{tab:table_5}} and \figurename{~\ref{fig:fig10}}), we have the following observations.
First, compared to \textbf{model A}, \textbf{model B} achieves a consistently superior accuracies (TAR and TPIR) by 1.3-4.1\% for TAR at FAR = 0.00001-0.01 in the verification task, 3.4-6.0\% for TPIR at FPIR = 0.01 and 0.1 in the identification open set task, and 1.2\% for Rank-1 in the identification close set task.
Second, \textbf{model C} shows a consistently higher accuracy than \textbf{model A} by the improvement of 2.6-9.8\% TAR at FAR = 0.001-0.1 in the verification task, 8.6-12.1\% TPIR at FPIR = 0.01-0.1 in the identification open set task, and 6.2\% Rank-1 in the identification close set task.
Third, \textbf{model C} shows a consistently higher accuracy than \textbf{model B} by the improvement of 1.3-6.5\% TAR at FAR = 0.001-0.1 in the verification set task, 5.2-6.1\% TPIR at FPIR = 0.01-0.1 in the identification open set task, and 5.0\% for Rank-1 in the identification close set task.
Last, although \textbf{model C} is trained from scratch, it outperformed the state-of-the-art method (Comparator Net \cite{ComparatorNet_ECCV2018}) by 0.4-3.6\% at FAR = 0.0001-0.01 in verification task, another state-of the-art method (PRN${}^{+}$ \cite{PRN_FR_ECCV2018}) by 3.8\% Rank-1 of identification close set task, and 5.0\% TPIR at FPIR = 0.01 in the identification open set task on the IJB-B dataset.
This validates the effectiveness of the proposed AFRN with the pair selection on the large-scale and challenging unconstrained face recognition.

\noindent\textbf{More Experiments on the CALFW, CPLFW, CFP, AgeDB, and IJB-C datasets.}
Due to the limited space, we provide more experiments in Section A in the supplementary material.

\section{Conclusion}\label{sec:conclusion}
We proposed the Attentional Feature-pair Relation Network (AFRN) which represented the face by the relevant pairs of local appearance block features with their weighted attention scores.
The AFRN represented the face by all possible pairs of the 9$\times$9 local appearance block features and the importance of each pair is weighted by the attention map that was obtained from adopting the low-rank bilinear pooling. We selected top-$K$ block feature-pairs as relevant facial information, dropped the remaining irrelevant. The weighted pairs of local appearance block features were propagated to extract the joint feature-pair relation by using bilinear attention network.
In experiments, we showed that the proposed AFRN achieved new state-of-the-art results in the 1:1 face verification and 1:N face identification tasks compared to current state-of-the-art methods on the challenging LFW, YTF, CALFW, CPLFW, CFP, AgeDB, IJB-A, IJB-B, and IJB-C datasets.

%\section*{Acknowledgment}
\paragraph*{Acknowledgment.}
%This work was supported by the MSIT, Korea, under the SW Starlab support program (IITP-2017-0-00897), Development of an Open Dataset and Cognitive Processing Technology for the Recognition of Features Derived From Unstructured Human Motions Used in Self-driving Cars (IITP-2018-0-01290) supervised by the IITP, and StradVision, Inc.
This research was supported by the MSIT(Ministry of Science, ICT), Korea, under the SW Starlab support program (IITP-2017-0-00897) supervised by the IITP (Institute for Information \& communications Technology Promotion), IITP grant funded by MSIT (IITP-2018-0-01290), and also supported by StradVision, Inc.

{\small
\bibliographystyle{ieee_fullname}
\bibliography{afrn_ref}
}

\clearpage
\appendix

\appendix
\section{More Experiments}
\subsection{Evaluation on the CALFW, CPLFW, CFP, and AgeDB datasets}\label{appendix:1}
We conduct experiments to demonstrate the effects of the proposed method on the Cross-Age LFW (CALFW) \cite{CALFW}, Cross-Pose LFW (CPLFW) \cite{CPLFW}, Celebrities in Frontal-Profile in the Wild (CFP) \cite{CFP-FP}, and AgeDB \cite{AgeDB} datasets.
\\
\\
\noindent\textbf{CALFW.} 
The CALFW is constructed by reorganizing the LFW \cite{LFW,LFWTechUpdate} verification pairs with apparent age gaps as large as possible to form the positive pairs and then selecting negative pairs using individuals with the same race and gender. The CALFW is more challenging than LFW. Similar to LFW, CALFW evaluation consists of verifying 6,000 pairs of images in 10 folds and report the average accuracy.
\\
\\
\noindent\textbf{CPLFW.} 
%The CPLFW is also a renovation of LFW. It deliberately searches and selects 3,000 positive face pairs with pose difference to add pose variation to intraclass variance. Negative pairs with same gender and race are also selected to reduce the influence of attribute difference between positive/negative pairs. The CPLFW dataset is more focused on cross-pose face recognition, and is more challenging than LFW.
The CPLFW is also constructed by reorganizing the LFW verification pairs by searching and selecting of 3,000 positive pairs with pose difference to add pose variation to intra-class variance. Negative pairs are also reorganized to reduce the influence of attribute differences between positive and negative pairs. Therefore, the CPLFW is more focused on cross-pose face recognition, and is more challenging than the LFW.
\\
\\
\noindent\textbf{CFP.} 
%The CFP consists of 500 subjects each with 10 frontal and 4 profile images. The evaluation protocol includes frontal-frontal (FF) and frontal-profile (FP) face verification, each having 10 folders with 350 same-person pairs and 350 different-person pairs.
The CFP consists of 500 subjects each with 10 frontal and 4 profile images. The evaluation protocol includes frontal-frontal (FF) and frontal-profile (FP) face verification, and each protocol has 10 folders with 350 positive pairs with same identity and 350 negative pairs with different identities.
\\
\\
\noindent\textbf{AgeDB.}
%The AgeDB is an in the-wild dataset with large variations in pose, expression, illuminations, and age. AgeDB contains 12,240 images of 440 distinct subjects. The minimum and maximum ages are 3 and 101, respectively. The test set is divided into four groups with different year gaps (5, 10, 20 and 30 years). Each group has ten split of face images, and each split contains 300 positive examples and 300 negative examples.
The AgeDB is a dataset for age invariant face recognition in the wild with in pose, expression, illumination, and age. The AgeDB contains 12,240 images of 440 unique subjects. The minimum and maximum ages are 3 and 101 years old, respectively. The test set is divided into four groups with different year gaps such as 5, 10, 20, and 30 years. Each group has ten split of face images, and each split includes 300 psotive examples and 300 negative examples.
\\
\\
\noindent\textbf{Evaluation Results.}
In image-based recognition on the CALFW, CPLFW, CFP, and AgeDB, we use a squared $L_{2}$ distance threshold to determine the classification of same and different.
\tablename{~\ref{tab:table_222} shows that our proposed AFRN with pair selection (\textbf{model C}) itself provides better accuracy than the AFRN without pair selection (\textbf{model B}). Finally, the \textbf{model C} acheives the outperformed accuracy and the \textit{state-of-the-art} results on the CALFW, CPLFW, CFP, AgeDB, respectively.

\begin{table}[t]
	\small
	\centering
	\caption{Performances of the proposed face recognition method on the CALFW, CPLFW, CFP, and AgeDB datasets.}
	\label{tab:table_222}
	\resizebox{\linewidth}{!} {
		\begin{tabular}{lcccc}
			\toprule
			\multicolumn{1}{c}{Method}   & CALFW & CPLFW & CFP & AgeDB  \\
			\hline
			CenterFace \cite{CenterFace} & $85.48$ & $77.48$  & - & - \\
			SphereFace \cite{SphereFace} & $90.30$ & $81.40$  & $94.38$ & $91.70$ \\
			VGGFace2 \cite{VGG2Face}     & $90.57$ & $84.00$  & - & - \\
			ArcFace \cite{ArcFace}       & $95.45$ & $92.08$  & $95.56$ & $95.15$ \\ \midrule
			\textbf{model B} (AFRN w/o pair selection) & $94.57$  & $91.17$ & $93.30$ & $93.40$ \\
			\textbf{model C} (AFRN w/ pair selection)  & $\mathbf{96.30}$  & $\mathbf{93.48}$  & $\mathbf{95.56}$ & $\mathbf{95.35}$ \\ \bottomrule
		\end{tabular}
	}
\end{table}

\begin{table*}[t]
	\caption{Comparison of performances of the proposed AFRN method with the \textit{state-of-the-art} on the IJB-C dataset. For verification, TAR \textsl{vs.} FAR are reported. For identification, TPIR \textsl{vs.} FPIR and the Rank-N accuracies are presented.}
	\label{tab:table_A2}
	\resizebox{\textwidth}{!} {
		\begin{tabular}{@{}lllllllllll@{}}
			\toprule
			\multicolumn{1}{c}{\multirow{2}{*}{Method}} & \multicolumn{4}{c}{1:1 Verification TAR} & \multirow{2}{*}{} & \multicolumn{5}{c}{1:N Identification TPIR}  \\ \cmidrule(lr){2-5} \cmidrule(l){7-11}
			\multicolumn{1}{c}{}                        	& FAR=0.00001  &  FAR=0.0001  &  FAR=0.001  &  FAR=0.01  &  &  FPIR=0.01  &  FPIR=0.1  &  Rank-1  &  Rank-5  &  Rank-10  \\ \midrule
			VGGFace2 \cite{VGG2Face}       	                & $0.747$  & $0.840$  & $0.910$  & $0.960$  & &  $0.746\pm0.018$  &  $0.842\pm0.022$  &  $0.912\pm0.017$  &  $0.949\pm0.010$  &  $0.962\pm0.007$ \\
			VGGFace2\_ft \cite{VGG2Face}   	                & $0.768$  & $0.862$  & $0.927$  & $0.967$  & &  $0.763\pm0.018$  &  $0.865\pm0.018$  &  $0.914\pm0.020$  &  $0.951\pm0.013$  &  $0.961\pm0.010$ \\
			CenterFace \cite{CenterFace}	                & $0.781$  & $0.853$  & $0.912$  & $0.953$  & &  $0.772\pm0.026$  &  $0.853\pm0.015$  &  $0.907\pm0.013$  &  $0.941\pm0.007$  &  $0.952\pm0.004$ \\
			Comparator Net \cite{ComparatorNet_ECCV2018}    & -        & $0.885$  & $0.947$  & $0.983$  & &  -  &  -  &  -  &  -  &  -   \\ 
			ArcFace \cite{ArcFace} 	                        & $0.883$  & $0.924$  & $0.956$  & $0.977$  & &  -  &  -  &  -  &  -  &  - \\ 
			Rajeev \textit{et. al} \cite{FaceDIV}           & $0.869$  & $0.925$  & $0.959$  & $0.979$  & & $0.873\pm0.032$ & $0.925\pm0.017$ & $0.949\pm0.018$ & $0.969\pm0.010$ & $0.975\pm0.009$ \\ \midrule
			\textbf{model A} (baseline)                     & $\mathbf{0.794}$  & $\mathbf{0.865}$  &  $\mathbf{0.921}$  &  $\mathbf{0.958}$ & &  $\mathbf{0.785\pm0.022}$  &  $\mathbf{0.870\pm0.021}$  &  $\mathbf{0.918\pm0.017}$  &  $\mathbf{0.949\pm0.013}$  &  $\mathbf{0.958\pm0.010}$  \\
			\textbf{model B} (AFRN w/o pair selection)      & $\mathbf{0.851}$  & $\mathbf{0.903}$  &  $\mathbf{0.951}$  &  $\mathbf{0.977}$ & &  $\mathbf{0.853\pm0.018}$  &  $\mathbf{0.905\pm0.018}$  &  $\mathbf{0.931\pm0.022}$  &  $\mathbf{0.956\pm0.010}$  &  $\mathbf{0.964\pm0.009}$  \\
			\textbf{model C} (AFRN w/ pair selection)      	& $\mathbf{0.883}$  & $\mathbf{0.930}$  &  $\mathbf{0.963}$  &  $\mathbf{0.987}$ & &  $\mathbf{0.884\pm0.017}$  &  $\mathbf{0.931\pm0.013}$  &  $\mathbf{0.957\pm0.015}$  &  $\mathbf{0.976\pm0.017}$  &  $\mathbf{0.977\pm0.007}$  \\ \bottomrule
		\end{tabular}
	}
\end{table*}

\subsection{Evaluation on the IJB-C dataset}
We also conduct experiments to demonstrate the effects of the proposed AFRN on the IJB-C \cite{IJB-C} datasets. 
%The IJB-C is a further extension of the IJB-B \cite{IJB-B_CVPRW2017}, which contains 3,531 unique subjects with a total of 31,334 still images (21,294 face and 10,040 non-face), averaging up to 6 images per subject, and 117,542 video frames collected in unconstrained settings, averaging up to 33 frames per subject and 3 videos per subject. Included with the protocols are two disjoint galleries, gallery 1 and gallery 2. Since the dataset contains two set of galleries 1 and 2, we report the average performance of both the gallery sets.
The IJB-C is an extenstion of the IJB-B, which contains a total of 31,334 still images with 3,531 unique subjects, and 117,542 video frames in unconstrained environments. It has an average of up to 6 imagew per subject, an average of up to 33 frames per subject and  3 videos per subject. Since the IJB-C contains two set of galleries 1 and 2, we report the average performance of both the gallery sets.

Three models (\textbf{model A}, \textbf{model B}, and \textbf{model C}) are trained on the roughly 2.8M refined VGGFace2 training set, with no people overlapping with subjects in the IJB-C dataset. For 1:1 face verification, we report the test results by using true accept rate (TAR) \textit{vs.} false accept rate (FAR) (\tablename{~\ref{tab:table_A2}}). For 1:N face identification, we report the results by using the true positive identification rate (TPIR) \textit{vs.} false positive identification rate (FPIR) and Rank-N (\tablename{~\ref{tab:table_A2}}). We average all the $1,024$ dimensional output vectors of the last fully connected layer of $\mathcal{F}_{\boldsymbol{\theta}}$ for a media in the template, then we average these media-averaged features to get the final template feature as face representation. Similarity to evaluation on the IJB-A and IJB-B, all performance evaluations are based on the squared $L_{2}$ distance threshold. 

\tablename{~\ref{tab:table_A2}} shows that the proposed \textbf{model C} shows a consistently higher accuracy than \textbf{model B} by the improvement of 1.0-3.2\% TAR at FAR = 0.00001-0.01 in the verification task, 2.6-3.1\% TPIR at FPIR = 0.01-0.1 in the identification open set task, and 2.6\% for rank-1 in the identification close set task. Although \textbf{model C} is trained from scratch, it outperformed the state-of-the-art method. This validates the effectiveness of the proposed AFRN with the pair selection on the large-scale and challenging unconstrained face recognition. 

From the experimental results (\tablename{~\ref{tab:table_A2}}), we have the following observations.
First, compared to \textbf{model A}, \textbf{model B} achieves a consistently superior accuracies (TAR and TPIR) by 1.9-5.7\% for TAR at FAR = 0.00001-0.01 in the verification task, 3.5-6.8\% for TPIR at FPIR = 0.01 and 0.1 in the identification open set task, and 1.3\% for Rank-1 in the identification close set task.
Second, \textbf{model C} shows a consistently higher accuracy than \textbf{model A} by the improvement of 2.9-8.9\% TAR at FAR = 0.00001-0.01 in the verification task, 6.1-9.9\% TPIR at FPIR = 0.01-0.1 in the identification open set task, and 3.9\% Rank-1 in the identification close set task.
Third, \textbf{model C} shows a consistently higher accuracy than \textbf{model B} by the improvement of 1.0-3.2\% TAR at FAR = 0.00001-0.01 in the verification set task, 2.6-3.1\% TPIR at FPIR = 0.01-0.1 in the identification open set task, and 2.6\% for Rank-1 in the identification close set task.
Last, although \textbf{model C} is trained from scratch, it outperformed the state-of-the-art method (Rajeev \textit{et. al} \cite{FaceDIV}) by 0.4-1.4\% at FAR = 0.00001-0.01 in verification task, 0.6-1.1\% TPIR at FPIR = 0.01-0.1 in the identification open set task, and 0.8\% Rank-1 of identification close set task on the IJB-C dataset.

The method proposed by Rajeev \textit{et. al} \cite{FaceDIV} is a fusion of ResNet-101 \cite{ResNet} and Inception ResNet-v2 \cite{Inception-v4} models. The Inception ResNet-v2 network has 224 \textit{conv.} layers, which are considerably more complex than our proposed AFRN method, and they used the training set with 5.6M images of 58,000 identities whereas we have a smaller number of subjects with 2.8M images of 8,900 identities. In order to obtain the comparable or better performance, it is considered that the proposed attention module and pair selection is effective because it obtains high performance even if a lesser amount of training images is used. This validates the effectiveness of the proposed AFRN with the pair selection on the large-scale and challenging unconstrained face recognition.

\end{document}